\documentclass{article}
\usepackage{ijcai24}

\usepackage{times}
\usepackage{soul}
\usepackage{url}
\usepackage[hidelinks]{hyperref}
\usepackage[utf8]{inputenc}
\usepackage[small]{caption}
\usepackage{graphicx}
\usepackage{amsmath}
\usepackage{amsthm}
\usepackage{booktabs}
\usepackage{algorithm}
\usepackage{algorithmic}
\usepackage[switch]{lineno}
\usepackage{amsfonts} 
\usepackage{utfsym}
\usepackage{newfloat}
\usepackage{listings}
\usepackage{tabularx}
\usepackage{multirow}
\usepackage{multicol}
\usepackage{makecell}
\usepackage{subfig}
\usepackage{array}
\usepackage{color}
\usepackage{amssymb}

\title{Fourier-enhanced Implicit Neural Fusion Network for Multispectral and Hyperspectral Image Fusion}

\author{
Yu-Jie Liang
\and
Zihan Cao\and
Shangqi Deng \And
Liang-Jian Deng\footnote{Corresponding author.}\\
\affiliations
UESTC
\emails
yujieliang0219@gmail.com,
iamzihan@std.uestc.edu.cn,
dengsq5856@126.com,
liangjian.deng@uestc.edu.cn
}

\begin{document}
\maketitle
\begin{abstract}

Recently, implicit neural representations (INR) have made significant strides in various vision-related domains, providing a novel solution for Multispectral and Hyperspectral Image Fusion (MHIF) tasks. However, INR is prone to losing high-frequency information and is confined to the lack of global perceptual capabilities. To address these issues, this paper introduces a Fourier-enhanced Implicit Neural Fusion Network (FeINFN) specifically designed for MHIF task, targeting the following phenomena: \textit{The Fourier amplitudes of the HR-HSI latent code and LR-HSI are remarkably similar; however, their phases exhibit different patterns.} In FeINFN, we innovatively propose a spatial and frequency implicit fusion function (Spa-Fre IFF), helping INR capture high-frequency information and expanding the receptive field. Besides, a new decoder employing a complex Gabor wavelet activation function, called Spatial-Frequency Interactive Decoder (SFID), is invented to enhance the interaction of INR features. Especially, we further theoretically prove that the Gabor wavelet activation possesses a time-frequency tightness property that favors learning the optimal bandwidths in the decoder. Experiments on two benchmark MHIF datasets verify the state-of-the-art (SOTA) performance of the proposed method, both visually and quantitatively. Also, ablation studies demonstrate the mentioned contributions. The code will be available on \href{https://anonymous.4open.science/r/FeINFN-15C9/}{Anonymous GitHub} after possible acceptance.
\end{abstract}

\section{Introduction}
\begin{figure}[!ht]
    \centering
    \includegraphics[width=7.5cm]{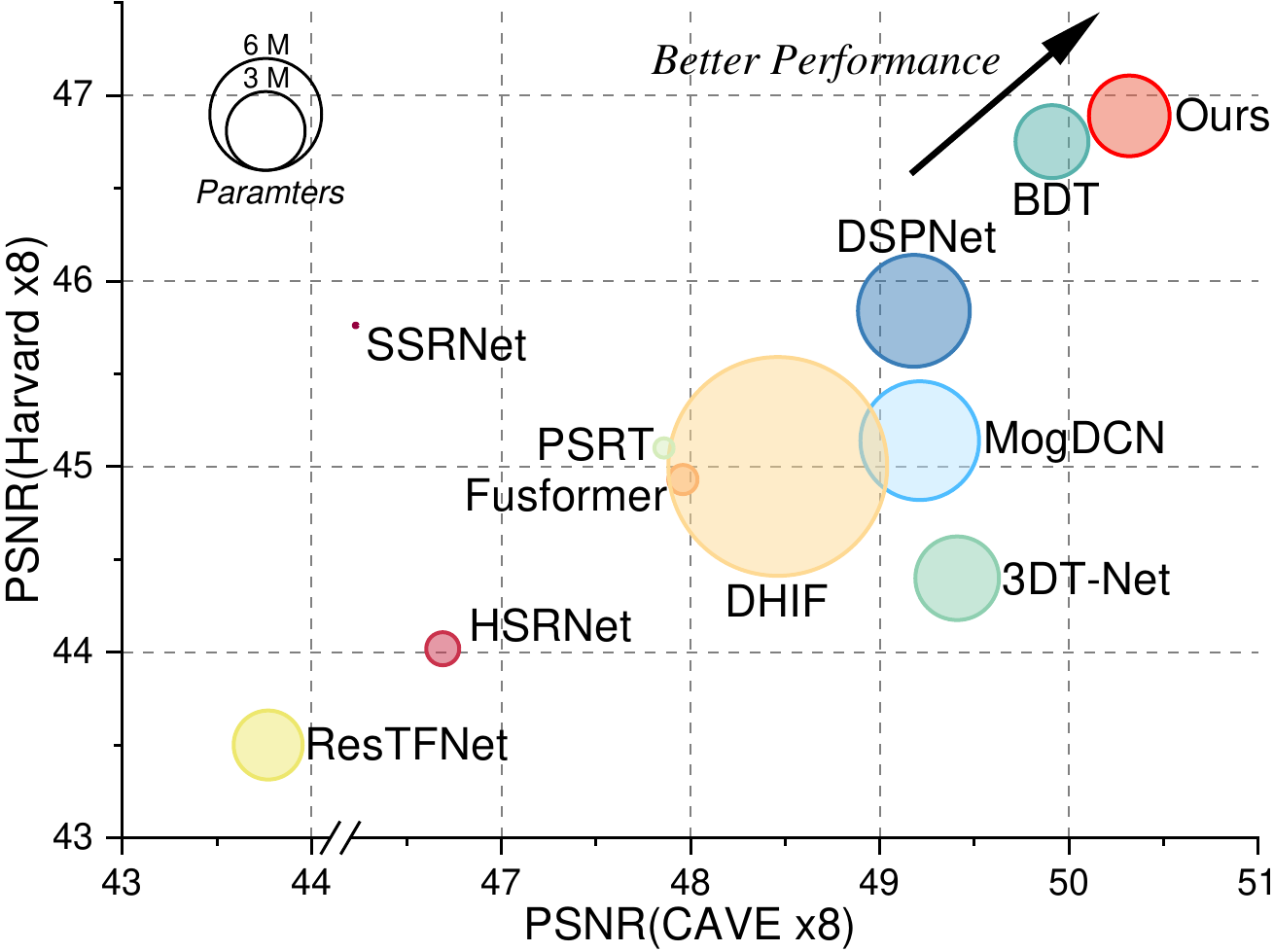}
    \caption{
    Comparison of our method with other methods on the CAVE($\times$ 8) and Harvard($\times$ 8) datasets. Closer to the top-right corner indicates better performance and the size of the circle indicates the number of parameters in the model.\vspace{-0.5cm}
    }
    \label{fig:psnr-comp}
\end{figure}

Hyperspectral imaging captures scenes across contiguous spectral bands, offering intricate details compared to traditional single or limited-band images, and 
improving computer vision application accuracy, such as target recognition, classification, tracking, and segmentation~\cite{fauvel2012advances,uzair2013hyperspectral,van2010tracking,Tarabalka2023Segmentation}. However, practical optical sensors face challenges in balancing spatial resolution and spectral precision. Images with over 100 bands often exhibit lower spatial resolution, while those with fewer bands display higher spatial resolution.
Efforts for MHIF are underway to fuse high spatial-resolution multispectral images (HR-MSI) with low spatial-resolution hyperspectral images (LR-HSI) to finally obtain high spatial-resolution hyperspectral images (HR-HSI). Actually, MHIF technology could fuse hyperspectral images with multispectral images, extracting information not detectable by HR-MSI to enhance richness and precision. Recent MHIF literature explores model-based approaches~\cite{dian2019hyperspectral,dian2019learning,xutgrs2022} and deep learning methods~\cite{huang2022deep,dong2021model,Cao2024DDIF}.
While model-based methods leverage image priors, challenges persist in obtaining high-fidelity, low-distortion HR-HSI due to the lack of large-scale training datasets. Among deep-learning approaches, CNN-based networks for HR-MSI and LR-HSI tend to be limited and lack interpretability for MHIF tasks and Transformer frameworks~\cite{hu2022fusformer,deng2023psrt} address the small receptive field of CNN but bring greater computational overhead.

In recent years, implicit representations of 3D scenes have garnered significant attention from researchers. For instance, Neural Radiance Field~\cite{wang2021nerf}  models 3D static scenes by mapping coordinates to signals through a neural network. Inspired by this, researchers have revisited image representation for 2D tasks. Recent studies~\cite{chen2021learning,lee2022local,sitzmann2020implicit,chen2023cascaded} have achieved arbitrary-scale super-resolution (SR) by replacing commonly used upsampling layers with local implicit image functions. Though these methods demonstrate superior performance in 2D tasks, they still have some drawbacks. \textit{Firstly}, INR calculates the RGB values of a queried coordinate based on the relative distances to the surrounding four pixels, treating it as a local operation in space that lacks consideration for global information. \textit{Additionally}, the MLP-ReLU structure used in traditional INR inherent high-frequency information bias~\cite{rahaman2019spectral} which is challenging to be eliminated during training.

To address these issues, we propose implicit fusion functions tailored for the MHIF task as a novel fusion paradigm. We first employ encoders to extract prior information from LR-HSI and HR-MSI, which is then fed into the implicit fusion functions in the form of latent codes. Unlike traditional INR, we transform latent codes into the Fourier domain and simultaneously perform spatial and frequency fusion in a unified network. This approach not only rectifies the high-frequency insensitivity induced by the MLP but also effectively extends the receptive field, encompassing a more comprehensive scope of global information. To integrate spatial and frequency domain representations efficiently, we design a decoder with time-frequency tightness, mapping features on both domains to pixel space. The contributions of this work are three folds:
\begin{itemize}
    \item We define a novel fusion framework based on INR, which innovatively extracts information from the spatial and  Fourier domains, effectively enhances the representation ability of high-frequency information, and expands the receptive field.
    \item We propose a new decoder employing a Gabor wavelet activation function to enhance the interaction of INR features. Furthermore, we theoretically prove that the complex Gabor wavelet activation possesses a time-frequency tightness property, which facilitates the decoder in learning the optimal bandwidths.
    \item The proposed network reaches state-of-the-art (SOTA) performance on the MHIF task across two widely used hyperspectral datasets at various fusion ratios. Fig.~\ref{fig:psnr-comp} provides a fair comparison with other SOTA methods.
\end{itemize}

\section{Related works}

\subsection{Implicit Neural Representation (INR)}
Unlike traditional discrete representations, neural implicit representation (INR) provides a more elegant and continuous parameterized approach. Initially applied in 3D modeling tasks, NeRF~\cite{wang2021nerf} revolutionized 3D computer vision by representing intricate three-dimensional scenes with just 2D pose images.
This line of work extends to the 2D imaging domain, where INR performs a weighted average on adjacent sub-codes to ensure output value continuity. LIIF~\cite{chen2021learning} recently introduces a local implicit image function for SR, leveraging MLP to sample pixel signals across the spatial domain. Several improvements focus on decoding networks; for example, UltraSR~\cite{xu2021ultrasr} incorporates residual networks, merging spatial coordinates and depth encoding. DIINN~\cite{nguyen2023single} utilizes a dual-interactive implicit neural network to decouple content and position features, improving decoding capabilities. JIIF~\cite{tang2021joint} proposes joint implicit image functions for multimodal learning, extracting priors from guided images.
Regarding activation functions in the MLP,  SIREN~\cite{sitzmann2020implicit} recommends utilizing periodic activation functions for continuous INR to fit complex signals.
On the other hand, WIRE~\cite{saragadam2023wire} further employs continuous complex Gabor wavelet activation functions to activate non-linearity, focusing more on spatial frequencies. 
However, there is limited research dedicated to designing INR architectures specifically for the MHIF task. The unique characteristics of hyperspectral images pose challenges for INR networks, in their insensitivity to high-frequency information.

\subsection{Latent Enhancement by Fourier Transform}
Fourier transform is a commonly used time-frequency analysis technique in signal processing, which converts signals from the time domain to the frequency domain. The Fourier domain has global statistical properties, and in recent years, many researchers have focused on processing frequency domain information in Fourier space.  Many works use the Fourier transform to enhance the representation ability of neural networks. For example, FDA~\cite{zhao2023wavelet} proposes exchanging amplitude and phase components in Fourier space between images to enhance and adjust frequency information. FFC~\cite{chi2020fast} introduces a novel convolution module that internally fuses cross-scale information to capture global features in Fourier space.
Similarly, GFNet~\cite{rao2021global} uses 2D discrete Fourier transform and inverse transform to extract features, implements learnable global filtering, and replaces the self-attention layer in Transformer.
UHDFour~\cite{Li2023ICLR} embeds Fourier transform into the image enhancement network to model global information. Together, these studies demonstrate the utility of frequency domain information in improving performance on visual tasks. We exploit the architecture of FeINFN to transform latent codes into the frequency domain, implicitly integrating representations of amplitude and phase components, and enhancing high-frequency injection.

\subsection{Motivation}\label{sec:2.4}
\begin{figure}
    \centering
    \includegraphics[width=8.5cm]{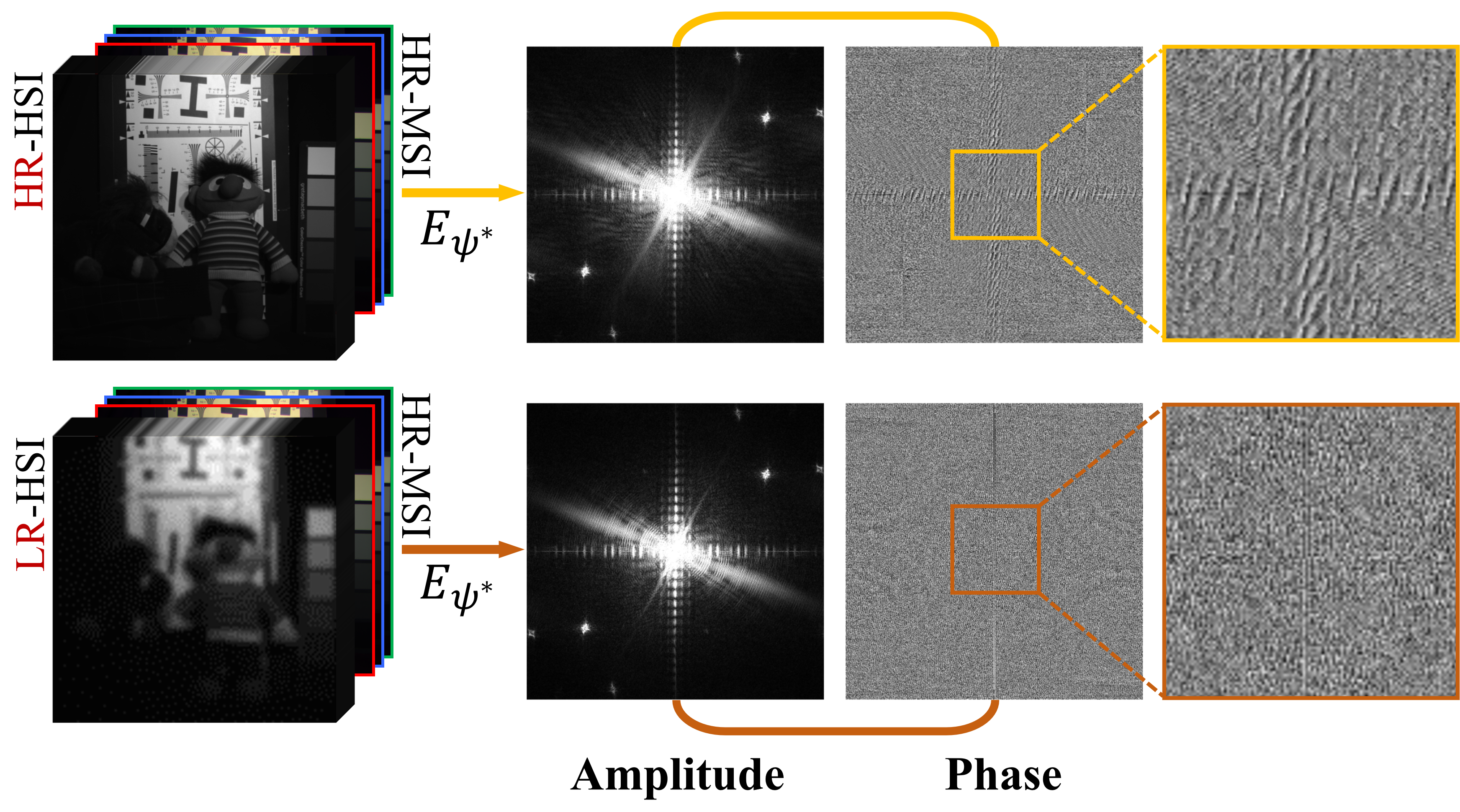}
    \caption{The amplitude of latent code from the encoder fed by HR-HSI and LR-HSI (combined with HR-MSI) share a similarity, but the phases differ from each other. $E_{\psi^*}$ is a trained encoder.\vspace{-0.4cm}}
    \label{amp-comp}
\end{figure}

\cite{rahaman2019spectral} finds that most neural networks exhibit a phenomenon of spectral bias through Fourier analysis. This includes neural networks such as MLP, which tend to learn low-frequency information during the early stages of training and are insensitive to high-frequency information. Moreover, we found this issue occurs in the MHIF task according to an experimental analysis as shown in Fig.~\ref{amp-comp}, where HR-HSI and LR-HSI were concatenated with HR-MSI and fed into a trained encoder to obtain latent codes. These codes were transformed into the frequency domain to visualize the amplitude and phase. 
It can be observed that the amplitudes from HR-HSI and LR-HSI are very similar, while the phases differ significantly. The phase of HR-HSI should naturally contain more texture than LR-HSI, a hypothesis validated by the visualized phase maps. Based on this finding, we transformed the latent codes into the Fourier domain to separately process amplitude and phase, to enhance the global learning of high-frequency information in the images. 

\section{Methodology}
In this section, we first present the preliminary of INR  and then provide the proposed framework tailored for MHIF task. Subsequently, we elaborate on the implementation details of the composited modules of the proposed FeINFN.

\subsection{Preliminary: Implicit Neural Representation}
Neural Radiance Fields~\cite{wang2021nerf} is represented by integral construction scenes. The value of a pixel in a certain viewing angle image is regarded as the integral of the characteristics of the sampling point from the proximal end to the far end of the ray. During actual training, the integral needs to be discretized. Extended to 2D image representation~\cite{chen2021learning}, it is sampled pixel by pixel from the vicinity of the query target.
Taking the low-resolution (LR) image $\mathbf I\in \mathbb{R}^{h \times w \times 3}$ upsampling to the high-resolution (HR) image ${\hat{\mathbf I}}\in \mathbb{R}^{H \times W \times 3}$ as an example, the process of generating the RGB values of the target coordinates $\mathbf{x}_q\in \mathbb{R}^2$ can be regarded as interpolation form, expressed as:

\begin{equation}
\hat{\mathbf I}(\mathbf{x}_q)=\sum_{i\in\mathcal{N}_q}w_{q,i}\mathbf{v}_{q,i},
\label{eq:2}
\end{equation}
where $\mathbf{v}_{q,i}\in \mathbb{R}^{4 \times 4 \times 3}$ is the interpolation pixel of $i$ interpolated by $q$'s surrounding pixels $\mathcal N_q\in \mathbb{R}^{4}$ and $w_{q,i}\in \mathbb{R}$ signifies the interpolation weight. In the implicit representation of local image features, the weights $w_{q, i}=S_i / S$, where $S_i$ represents the area formed by $q$ and $i$ in the diagonal region and $S$ denotes the total area enclosed by the set $\mathcal{N}_q$.

The interpolation value $\mathbf{v}_{q,i}$ is effectively generated by a basis function:
\begin{equation}
\mathbf{v}_{q,i} = \phi_{\theta}(\mathbf{z}_i,\mathbf{x}_q-\mathbf{x}_i),
\end{equation}
where $\phi_{\theta}$ is typically an MLP, $\mathbf{z}_i$ is the latent code generated by an encoder for the coordinates $\mathbf{x}_i$, and $\mathbf{x}_q-\mathbf{x}_i$ represents the relative coordinates. 
From the above equations, it can be inferred that the interpolation features can be represented by a set of local feature vectors in the LR domain.
Typically, interpolation-based methods~\cite{press2007numerical,keys1981cubic} achieve upsampling by querying $\mathbf{x}_q-\mathbf{x}_i$ in the arbitrary SR task. See more details in~\cite{chen2021learning}.

\begin{figure*}[!ht]
    \centering
    \includegraphics[width=\textwidth]{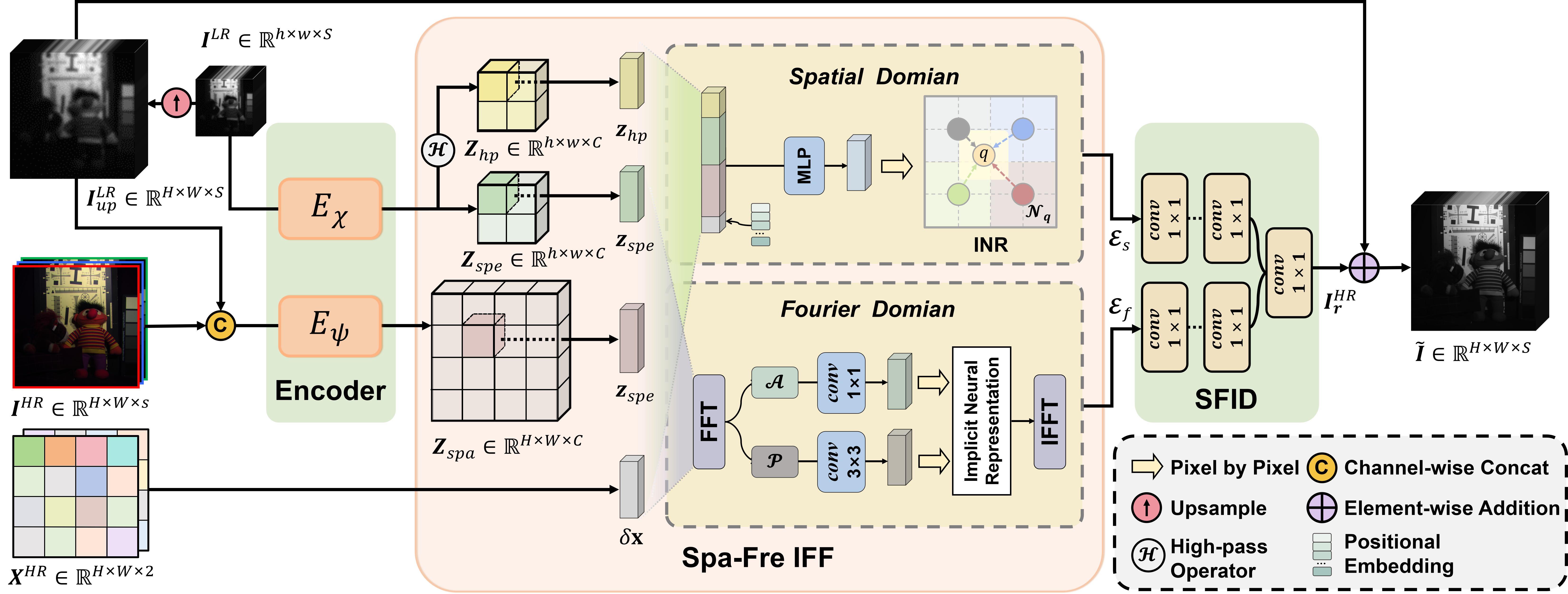}
    \caption{The flowchart of the FeINFN framework which is composed of a spectral encoder $E_{\chi}$, a spatial encoder $E_{\psi}$, MHIF task-designed spatial and Fourier domains implicit fusion functions, and a pixel space mapping decoder. Please note that $\mathbf I^{LR}$ is the LR-HSI, $\mathbf I^{HR}$ is the HR-MSI, $\mathbf I^{LR}_{up}$ is the bicubic interpolation LR-HSI, and $\mathbf X^{HR}$ is the HR normalized 2D coordinate map.  $\mathbf z_{spe}$, $\mathbf z_{spa}$, $\mathbf z_{hp}$, $\delta \mathbf x$ correspond to individual pixel units, $\mathcal{A}$ and $\mathcal{P}$ represents amplitude and phase, respectively.\vspace{-0.5cm}}
    \label{fig:2}
\end{figure*}

\subsection{Overview of the FeINFN Framework}
In this work, we propose the FeINFN, which adopts a novel framework for simultaneously performing neural implicit representation in both the spatial and frequency domains to execute the MHIF task. Fig.~\ref{fig:2} provides an overview of the proposed framework, designed to fuse LR-HSI $\mathbf I^{LR} \in \mathbb{R}^{h \times w \times S}$ and HR-MSI $\mathbf I^{HR} \in \mathbb{R}^{H \times W \times s}$ to generate HR-HSI images $\widetilde{\mathbf I} \in \mathbb{R}^{H \times W \times S}$ based on a given upsampling scale $r$. 

Initially, the LR-HSI is fed into encoder $E_\chi$ to extract spectral features  $\mathbf Z_{spe} \in \mathbb{R}^{h \times w \times C}$. Simultaneously, the concatenated bicubic interpolation LR-HSI $\mathbf I_{up}^{LR} \in \mathbb{R}^{H \times W \times S}$ and $\mathbf I^{HR}$, are fed into encoder $E_\psi$ to extract spatial features $\mathbf Z_{spa} \in \mathbb{R}^{H \times W \times C}$. Additionally, the pixel's central position is represented as the coordinate point. The coordinate map is normalized into a two-dimensional grid $[-1,1]\times [-1, 1]$, obtaining a HR normalized 2D coordinate map $\mathbf X^{HR} \in \mathbb{R}^{H \times W \times 2}$. The extracted $\mathbf Z_{spe}$ and $\mathbf Z_{spa}$, along with the 2D coordinates of $\mathbf I^{HR}$, are forwarded to Spatial-Frequency Implicit Fusion Function (Spa-Fre IFF), outputting spatial domain features $\mathcal {E}_{s} \in \mathbb{R}^{H \times W \times S}$ and frequency domain features $\mathcal E_{f} \in \mathbb{R}^{H \times W \times S}$. The $\mathcal E_{s}$ and $\mathcal E_{f}$ serve as inputs to a pixel space mapping decoder which generates the residual image $\mathbf I_r^{HR} \in \mathbb{R}^{H \times W \times S}$. Finally, the residual image $\mathbf I_r^{HR}$ is combined with the bicubicly upsampled image $\mathbf I_{up}^{LR}$ via element-wise addition, yielding the ultimate fusion image $\widetilde{\mathbf I}$.

\subsection{INR Encoder Networks}  
Analogous to local implicit representation functions~\cite{chen2021learning,lee2022local,sitzmann2020implicit,chen2023cascaded}, the initial step involves extracting latent code representations.
For the MHIF task, we address the challenges of both upsampling and fusion simultaneously, employing implicit neural representations as the solution.
The INR encoders try to extract spatial and spectral latent codes $\mathbf Z_{spa}\in \mathbb{R}^{H \times W \times C}, \mathbf Z_{spe}\in\mathbb{R}^{h \times w \times C}$:
one is extracted from $\mathbf I^{LR}$, serving as the carrier for spectral information; the other is encoded from the concatenation of $\mathbf I_{up}^{LR}$ and $\mathbf I^{HR}$, aiding in spatial information during the fusion process. This process can be denoted as:
\begin{equation}
\begin{cases}
\mathbf Z_{spe}=E_{\chi}(\mathbf I^{LR}),\\
\mathbf Z_{spa}=E_{\psi}\left({\rm{Cat}}(\mathbf I_{up}^{LR},\mathbf I^{HR})\right),
\end{cases}
\end{equation}
where $E_{\chi}$ is the spectral encoder parameterized by $\chi$, $E_{\psi}$ is the spatial encoder parameterized by $\psi$, and ${\rm{Cat}}(\mathbf I_{up}^{LR}, \mathbf I^{HR})$ denotes the concatenation along the channel dimension. In practice, we utilize  EDSR~\cite{lim2017enhanced} as INR encoder networks.

\subsection{Spatial-Frequency Implicit Fusion Function}
To address the mentioned issues~\ref{sec:2.4}, we propose Spatial-Frequency Implicit Fusion Function, dubbed Spa-Fre IFF which is a dual-branch fusion function and utilized for computing the fusion feature of $\mathbf Z_{spe}$ and $\mathbf Z_{spa}$ in the spatial and frequency domains, respectively. Given a queried HR coordinate $\mathbf {x}_{q}\in\mathbf{X}^{HR}$ of a pixel unit $q$, Spa-Fre IFF estimates spatial feature vector ${\boldsymbol\varepsilon}_{s}\in\mathbb{R}^{1\times1\times S}$ ($\boldsymbol\varepsilon_{s} \in\mathcal E_{s}$) and frequency feature vector ${\boldsymbol\varepsilon}_{f}\in\mathbb{R}^{1\times1\times S}$ ($\boldsymbol\varepsilon_{f} \in\mathcal E_{f}$)as follows:
\begin{equation}
{\boldsymbol\varepsilon}_{s},{\boldsymbol\varepsilon}_{f}=\text{Spa-Fre IFF}(\mathbf z_{spe},\mathbf z_{spa},\delta{\mathbf x}),
\end{equation}
where $\mathbf z_{spe}\in\mathbb{R}^{1\times1\times C}$ represents the spectral latent code vector corresponding to $\mathbf x_{q}$, and $\mathbf z_{spa}\in\mathbb{R}^{4\times4\times C}$ 
 is the spatial latent code vector. $\delta{\mathbf x}$ denotes the set of local relative coordinates, expressed by the following formula:
\begin{equation}
\delta {\mathbf x}=\left\{\mathbf x_q-\mathbf x_{q,i}\right\}_{i\in\mathcal{N}_q},
\end{equation}
where $\mathbf x_{q,i}$ refers to the coordinates most proximate to the query coordinate $\mathbf x_{q}$, representing the four corner pixels closest to $q$ in the HR space.

\noindent\textbf{Spatial Implicit Fusion Function: }The Spatial Implicit Fusion Function aims to leverage the powerful representation capabilities of INR to achieve implicit fusion in the spatial domain, as shown in Fig.~\ref{fig:2} (see branch ``Spatial Domain''). Specifically, we employ high-pass operators $\mathcal{H}$ to filter the spectral latent codes, 
as a complement to the high-frequency information on the spectrum:
\begin{equation}
    \mathbf z_{hp}=\mathcal{H}(\mathbf z_{spe}) ,
\end{equation}
where $\mathbf z_{hp}\in\mathbb{R}^{1\times1\times C}$ represents the high-frequency latent code of $\mathbf I^{LR}$.
Also, we suggest frequency encoding for relative positional coordinates as follows:
\begin{equation}
\begin{aligned}
        \gamma(\delta \mathbf x)=&[\sin(2^{0}\delta \mathbf x),\cos(2^{0}\delta \mathbf x),\cdots,
    \\&\sin(2^{L-1}\delta \mathbf x),\cos(2^{L-1}\delta \mathbf x)] ,
\end{aligned}
\end{equation}
where $L$ is a hyperparameter, in practice, we set $L$ to 10.
Additionally, leveraging the graph attention mechanism~\cite{tang2021joint}, we parameterize the solution for interpolation weights $\mathbf w_{q, i} \in \mathbb{R}^{1 \times S}$, and the implicit fusion function simultaneously outputs fusion interpolation values $\mathbf v_{q, i} \in \mathbb{R}^{4 \times 4 \times S}$ and interpolation weights $\mathbf w_{q, i}$. The implicit fusion function is specifically expressed as:
\begin{equation}
\mathbf w_{q,i},\mathbf v_{q,i}=\phi_{\theta}(\mathbf z_{spe},\mathbf z_{spa},\mathbf z_{hp},\gamma(\delta \mathbf x)) ,
\end{equation}
where $\phi_{\theta}$ is an MLP parameterized by $\theta$.
The spatial implicit fusion interpolation, as shown in Eq.~\eqref{eq:2}, yields the fused spatial feature ${\boldsymbol\varepsilon}_s\in\mathbb{R}^{1\times1\times S}$ and can be described as follows:
\begin{equation}
{\boldsymbol\varepsilon}_s=\sum_{i\in\mathcal{N}_q}{\overline{\mathbf w}_{q,i}}* \mathbf v_{q,i} .
\end{equation}


\begin{figure}[t]
    \centering
    \includegraphics[width=6.5cm]{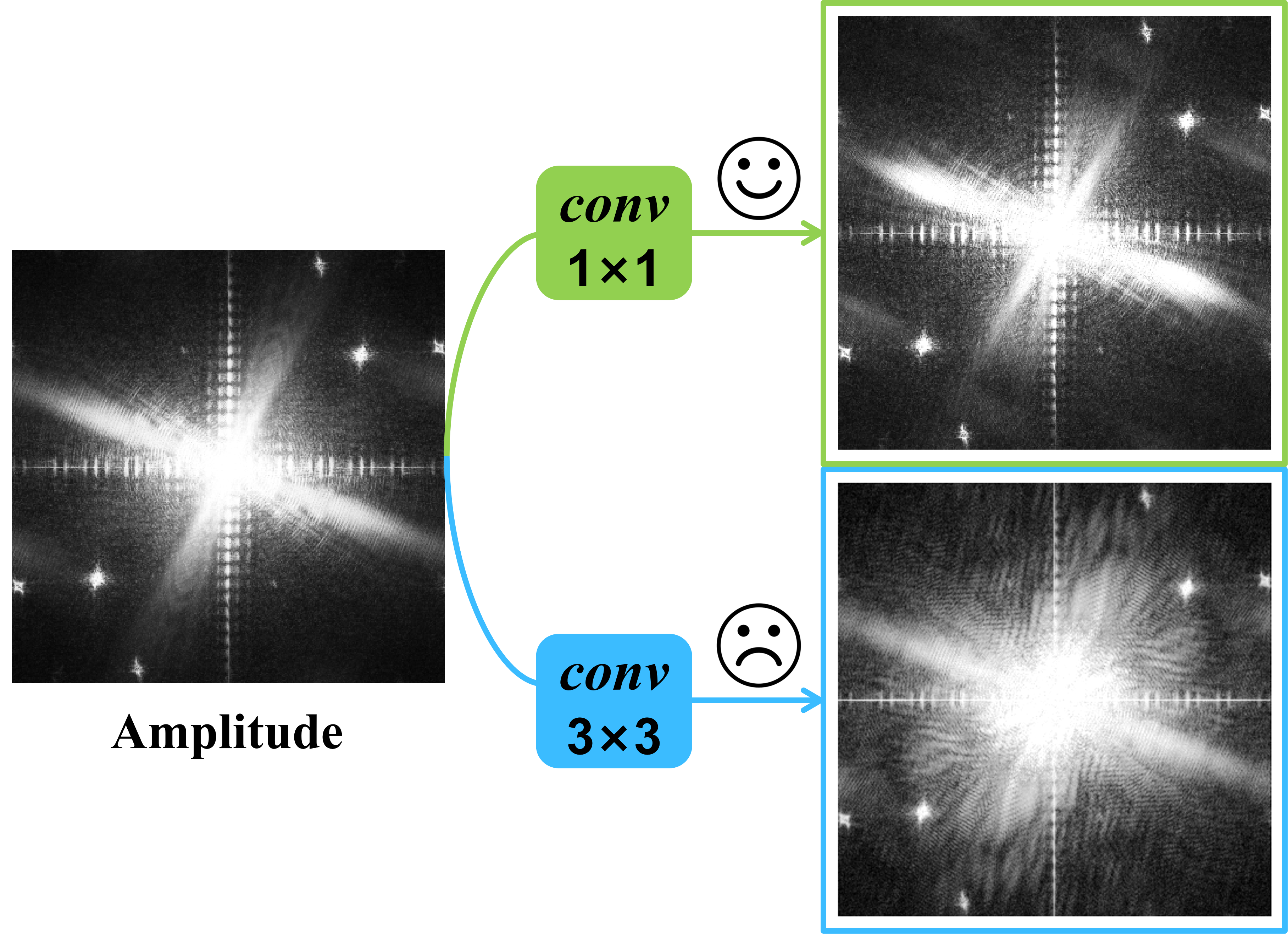}
    \caption{$3\times 3$ convolution would suffer from the issue of spectrum leakage, which can be alleviated by $1\times 1$ convolution.\vspace{-0.5cm}}
    \label{conv-comp}
\end{figure}
\noindent\textbf{Frequency Implicit Fusion Function: }From Fig.~\ref{amp-comp}, 
we observed characteristics in the frequency features between LR-HSI and HR-HSI. Hence, we design a frequency implicit fusion function to express global features continuously in the Fourier domain. Notably, directly applying static kernel convolution in the frequency domain would only enhance a specific frequency range, which is inappropriate for the fusion task. However, by learning feature content to generate weights, INR can be seen as a dynamic interpolation method in continuous space, adaptively enhancing information in the frequency domain without overly altering the frequency distribution. Therefore, introducing INR into the Fourier domain is reasonable. Since amplitude and phase exhibit different forms, as shown in Fig.~\ref{amp-comp}, we handle them separately.

With the considerations mentioned above, as illustrated in Fig.~\ref{fig:2} (see branch ``Fourier Domain''), we initially employ FFT to transform latent codes $\mathbf{z}_{spe}$ and $\mathbf{z}_{spa}$ from the spatial domain to the frequency domain, obtaining $\mathbf{f}_{spe}\in\mathbb{R}^{1\times1\times C}$ and $\mathbf{f}_{spa}\in\mathbb{R}^{4\times4\times C}$.
After the transformation, we further obtain amplitude components $\mathcal{A}(\mathbf{f}_{spe})$ and $\mathcal{A}(\mathbf{f}_{spa})$,  as well as phase components $\mathcal{P}(\mathbf{f}_{spe})$ and $\mathcal{P}(\mathbf{f}_{spa})$.

\textit{For the amplitude}, as shown in Fig.~\ref{conv-comp}, the amplitude distribution of LR-HSI and HR-HSI are very similar, and the non-point-wise convolution (\emph{e.g.} Conv $3\times 3$) causes an issue of spectrum leakage, confusing channel information. 
In contrast, point-wise convolution does not span multiple locations in the frequency domain and has no overlap allowing it to capture information across channels effectively.
Thus the fusion function for amplitude components is more suitable when applying point-wise convolution:
\begin{equation}
\mathbf w^\mathcal{A}_{q,i},\mathbf v^\mathcal{A}_{q,i}={\mathbf{\phi}_\alpha^\mathcal{A}}(\mathcal{A}(\mathbf{f}_{spe}),\mathcal{A}(\mathbf{f}_{spa}),\delta \mathbf x ) ,
\label{eq:16}
\end{equation}
where $\mathbf w^\mathcal{A}_{q, i}\in\mathbb{R}^{1 \times S}$ and $\mathbf v^\mathcal{A}_{q, i}\in\mathbb{R}^{4\times4\times S}$   are the weights and interpolated values for the corresponding amplitude component, and ${\mathbf{\phi}_\alpha^\mathcal{A}}$ is a simple network composed of two layers of point convolutions parameterized by $\alpha$.
Similar to operations in the spatial domain,
implicit fusion interpolation is performed after obtaining interpolated values $\mathbf{v}^\mathcal{A}_{q, i}$ and the normalized weights ${\overline{\mathbf w}^\mathcal{A}_{q,i}}$:
\begin{equation}
\mathcal{A}^{\prime}_f=\sum_{i\in\mathcal{N}_q}{\overline{\mathbf w}^\mathcal{A}_{q,i}}
*\mathbf v^\mathcal{A}_{q,i} ,
\label{eq:17}
\end{equation}
where $\mathcal{A}^{\prime}_f\in\mathbb{R}^{1\times1\times S}$ is the integrated amplitude component.

\textit{For the phase}, which encapsulates information such as texture details, LR-HSI and HR-HSI often have different phase information. It is known that point convolutions fail to capture sufficient spatial representations. Therefore, we use a $3\times 3$ convolution to learn phase information. Additionally, small changes in the frequency domain may result in significant variations in the spatial domain.
We still consider using the form of INR interpolation for phase learning. The handling of the phase components $\mathcal{P}(\mathbf{f}_{spe})$ and $\mathcal{P}(\mathbf{f}_{spa})$ are formally similar to Eqs.~\eqref{eq:16} and~\eqref{eq:17}:
\begin{align}
\mathbf{w}^\mathcal{P}_{q,i},\mathbf{v}^\mathcal{P}_{q,i}&={\mathbf{\phi}_\beta^\mathcal{P}}(\mathcal{P}(\mathbf{f}_{spe}),\mathcal{P}(\mathbf{f}_{spa}),\delta(\mathbf{x})) ,\\
\mathcal{P}^{\prime}_f&=\sum_{i\in\mathcal{N}_q}{\overline{\mathbf{w}}^\mathcal{P}_{q,i}}* \mathbf{v}^\mathcal{P}_{q,i}.
\end{align}
The simple network $\mathbf{\phi}_\beta^\mathcal{P}$ consists of two layers of $3\times 3$ convolutions parameterized by $\beta$. $\mathcal{P}^{\prime}_f\in\mathbb{R}^{1\times 1 \times S}$ represents the integrated phase component.

Finally, IFFT is applied to map the frequency features $\mathcal{A}^{\prime}_f$ and $\mathcal{P}^{\prime}_f$ back to the image space, obtaining the frequency domain feature $\varepsilon_{f}\in\mathcal E_{f}$. Since in frequency space, one frequency point may correspond to multiple pixels at different positions in the spatial domain, the receptive field of INR in the frequency domain is enlarged in the spatial domain.
\begin{figure}[t]
    \centering
    \includegraphics[width=8.5cm]{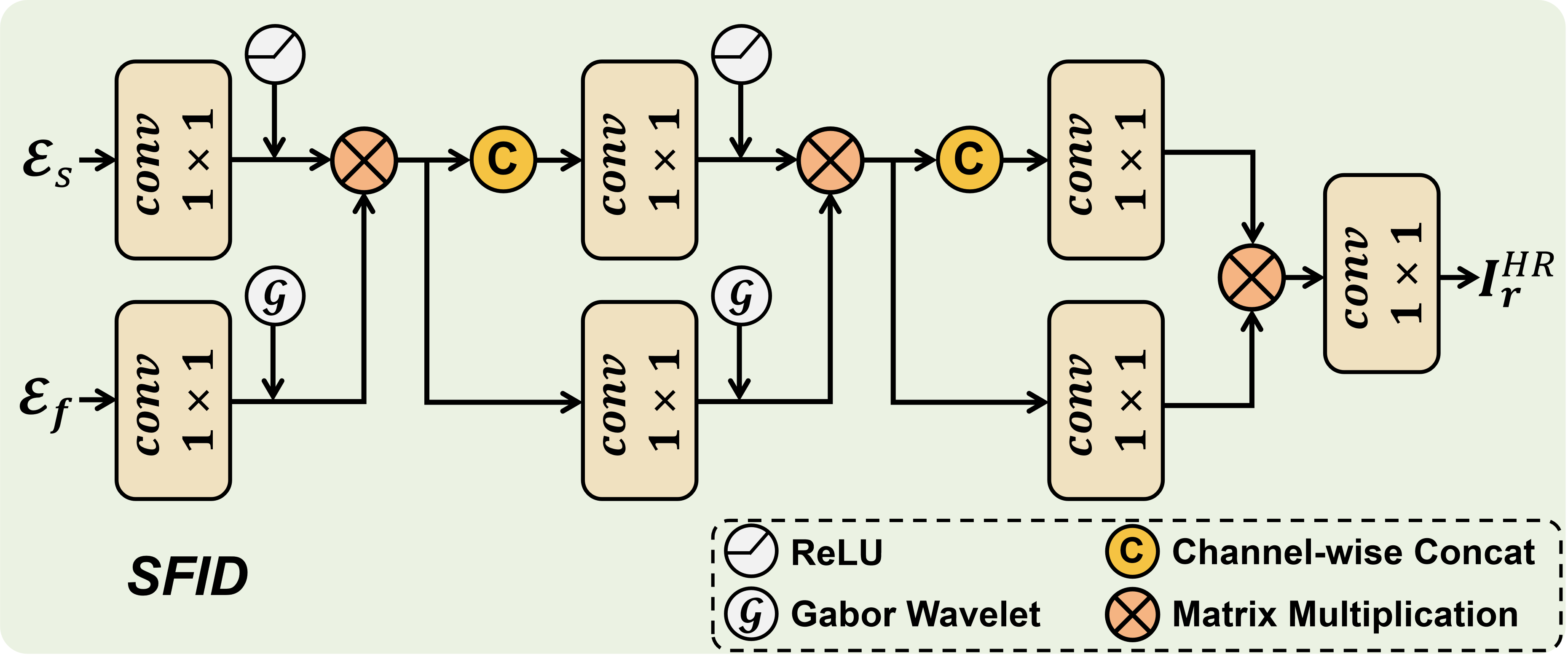}
    \caption{Detailed composition of the proposed SFID.\vspace{-0.4cm}}
    \label{fig:decoder}
\end{figure}

\begin{table*}[!t] 
		\setlength{\tabcolsep}{2pt}
    \renewcommand\arraystretch{1.05}
		\centering
		\resizebox{\linewidth}{!}{
			\begin{tabular}{@{}ccccccccccc@{}}
				\toprule
				\multirow{2}{*}{Methods} & \multicolumn{5}{c}{CAVE  $\times 4$} & \multicolumn{5}{c}{Harvard   $\times 4$} \\ \cmidrule(l){2-11} 
				&PSNR($\uparrow$) &SAM($\downarrow$) &ERGAS($\downarrow$) &SSIM($\uparrow$) &$\#$params &PSNR($\uparrow$) &SAM($\downarrow$) &ERGAS($\downarrow$) &SSIM($\uparrow$) &$\#$params  \\ \midrule
				
				Bicubic                  &34.33$\pm$3.88 &4.45$\pm$1.62 &7.21$\pm$4.90 &0.944$\pm$0.029 &\multicolumn{1}{c|}{\makecell[c]{$-$}} &38.71$\pm$4.33 &2.53$\pm$0.67 &4.45$\pm$1.81 &0.948$\pm$0.027 &\makecell[c]{$-$} \\
				
				
				CSTF-FUS~\cite{li2018fusing}&34.46$\pm$4.28 &14.37$\pm$5.30 &8.29$\pm$5.29 &0.866$\pm$0.075&\multicolumn{1}{c|}{\makecell[c]{$-$}}
				&39.15$\pm$3.45&6.93$\pm$2.69&4.66$\pm$1.81&0.914$\pm$0.049&\makecell[c]{$-$} \\
				
				LTTR~\cite{dian2019learning}&35.85$\pm$3.49 &6.99$\pm$2.55 &5.99$\pm$2.92 &0.956$\pm$0.029&\multicolumn{1}{c|}{\makecell[c]{$-$}}
				&40.88$\pm$3.94&4.01$\pm$1.27&4.03$\pm$2.18&0.957$\pm$0.035&\makecell[c]{$-$} \\
				
				LTMR~\cite{dian2019hyperspectral}&36.54$\pm$3.30 &6.71$\pm$2.19 &5.39$\pm$2.53&0.963$\pm$0.021&\multicolumn{1}{c|}{\makecell[c]{$-$}}
				&42.06$\pm$3.56&3.51$\pm$0.99&3.59$\pm$2.03&0.970$\pm$0.020&\makecell[c]{$-$} \\
				
				IR-TenSR~\cite{xutgrs2022}&35.61$\pm$3.45 &12.30$\pm$4.68 &5.90$\pm$3.05 &0.945$\pm$0.027 &\multicolumn{1}{c|}{\makecell[c]{$-$}}
				&40.47$\pm$3.04&4.36$\pm$1.52&5.57$\pm$1.57&0.963$\pm$0.014 &\makecell[c]{$-$} \\
				
				\hline
    
				ResTFNet~\cite{LIU20201}&45.58$\pm$5.47 &2.82$\pm$0.70 &2.36$\pm$2.59 &0.993$\pm$0.006 &
				\multicolumn{1}{c|}{\makecell[c]{2.387M}} &45.94$\pm$4.35 &2.61$\pm$0.69 &2.56$\pm$1.32 &0.985$\pm$0.008 &2.387M\\
				
				SSRNet~\cite{zhang2020ssr}&48.62$\pm$3.92 &2.54$\pm$0.84 &1.63$\pm$1.21 &0.995$\pm$0.002&
				\multicolumn{1}{c|}{\makecell[c]{\textbf{0.027M}}} &48.00$\pm$3.36 &2.31$\pm$0.60 &2.30$\pm$1.42 &0.987$\pm$0.007 &\textbf{0.027M}\\
				
				HSRNet~\cite{hu2022hyperspectral}&50.38$\pm$3.38 &2.23$\pm$0.66 &1.20$\pm$0.75 &0.996$\pm$0.001 &
				\multicolumn{1}{c|}{\makecell[c]{0.633M}} &48.29$\pm$3.03 &2.26$\pm$0.56 &1.87$\pm$0.81 &0.988$\pm$0.006 &0.633M\\
				
				MogDCN~\cite{dong2021model} &51.63$\pm$4.10 &2.03$\pm$0.62 &1.11$\pm$0.82 &0.997$\pm$0.002 &
				\multicolumn{1}{c|}{\makecell[c]{6.840M}}  &47.89$\pm$4.09 &2.11$\pm$0.52 &1.89$\pm$0.82
                & 0.988$\pm$0.007 &6.840M\\
				
				Fusformer~\cite{hu2022fusformer}&49.98$\pm$8.10 &2.20$\pm$0.85 &2.50$\pm$5.21 &0.994$\pm$0.011 &\multicolumn{1}{c|}{\makecell[c]{0.504M}}  &47.87$\pm$5.13 &2.84$\pm$2.07 &2.04$\pm$0.99 &0.986$\pm$0.010 &0.467M\\
				
				DHIF~\cite{huang2022deep} &51.07$\pm$4.17 &2.01$\pm$0.63 &1.22$\pm$0.97
                &0.997$\pm$0.002 &\multicolumn{1}{c|}{\makecell[c]{22.462M}}   
                & 47.68$\pm$3.85 &2.32$\pm$0.53 
                & 1.95$\pm$0.92
                & 0.988$\pm$0.007 &22.462M\\
				
				PSRT~\cite{deng2023psrt} &50.47$\pm$6.19 
                & 2.19$\pm$0.64 
                & 2.06$\pm$3.71
                & 0.996$\pm$0.003 &\multicolumn{1}{c|}{\makecell[c]{\underline{0.247M}}}   
                & 47.96$\pm$3.21 &2.18$\pm$0.55 
                & 1.89$\pm$0.86
                & 0.988$\pm$0.006 &{\underline{0.247M}}\\
				
				3DT-Net~\cite{ma2023learning} &51.38$\pm$4.18 &2.16$\pm$0.70 
                & 1.14$\pm$1.00
                & 0.996$\pm$0.003 &\multicolumn{1}{c|}{\makecell[c]{3.464M}}   
                & 47.78$\pm$4.42 &\textbf{2.04$\pm$0.51} & 1.98$\pm$0.86
        &\textbf{0.989$\pm$0.006} &3.464M\\
    
                DSPNet~\cite{sun2023dual} & 51.18$\pm$3.92 &2.15$\pm$0.64 & 1.13$\pm$0.82&0.997$\pm$0.002 &\multicolumn{1}{c|}{\makecell[c]{6.064M}}   & 48.29$\pm$3.16 &2.30$\pm$0.55 & 1.93$\pm$0.93& 0.988$\pm$0.006 &6.064M\\
                
			BDT~\cite{dengbidirectional} 
                &\underline{52.30$\pm$3.98} &\underline{1.93$\pm$0.55} &\underline{1.02$\pm$0.77} &\underline{0.997$\pm$0.001} &\multicolumn{1}{c|}{\makecell[c]{2.668 M}} &\underline{48.83$\pm$3.45} &\underline{2.07$\pm$0.49} &\underline{1.83$\pm$0.81} &0.989$\pm$0.007 &2.668 M\\
                
    
                 FeINFN(Ours)&
                 {\textbf{52.47$\pm$4.10}}&
                 {\textbf{1.91$\pm$0.59}}&
                 {\textbf{0.98$\pm$0.74}}&
                 {\textbf{0.998$\pm$0.002}}&
                 \multicolumn{1}{c|}{\makecell[c]{3.165 M}}&
                 {\textbf{49.06$\pm$3.15}}&
                 2.10$\pm$0.53&
                 {\textbf{1.78$\pm$0.75}}&
                 {\underline{0.989$\pm$0.007}}&
                 3.165 M\\
                 
				\bottomrule				
		\end{tabular}}
        \caption{The average and standard deviation calculated for all the compared approaches on 11 CAVE examples and 10 Harvard examples simulating a scaling factor of 4. The best results are in bold, second-best in underline. ``M'' refers to millions.\vspace{-0.5cm}}
		\label{table_x4}
	\end{table*}

\begin{figure*}[htbp]
    \centering
    \includegraphics[width=\textwidth]{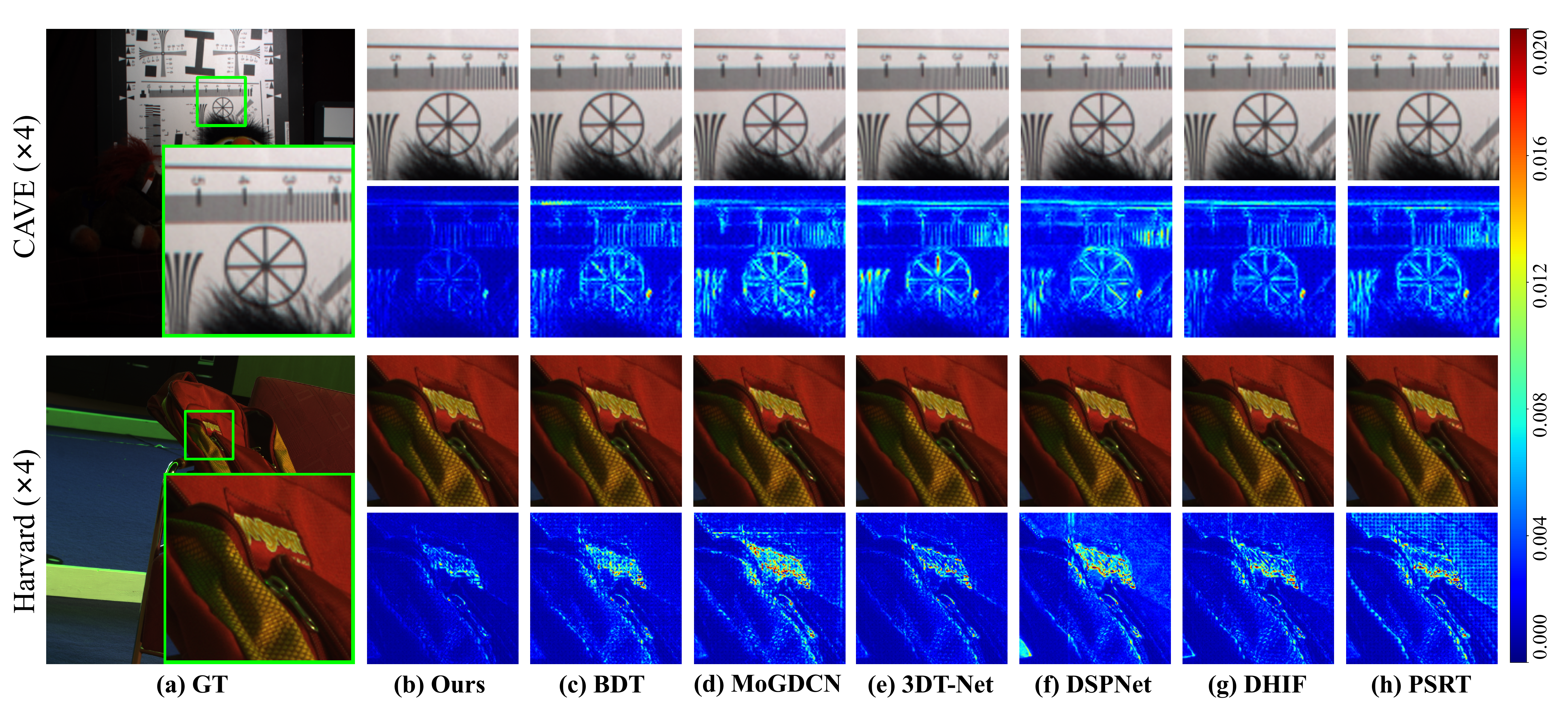}
    \caption{The upper and lower parts respectively showcase the results of ``\textit{Chart and Stuffed Toy}'' from the CAVE dataset and ``\textit{Backpack}'' from the Harvard dataset using pseudo-color representation. Green rectangles depict some close-up shots. The second and fourth rows show the residuals between the ground truth (GT) and the fusion products.}
    \vspace{-0.4cm}
    \label{fig:3}
\end{figure*}
\subsection{Spatial-Frequency Interactive Decoder}

After obtaining the spatial feature map and frequency domain feature map, it is essential to consider how to integrate them seamlessly. Firstly, our decoder needs to have dual input and interactive capabilities. Secondly, it is necessary to focus on representing images in the spatial-frequency domain. With this in mind, we introduce the complex Gabor wavelet activation function with good time-frequency tightness and propose the Spatial-Frequency Interactive Decoder (SFID). Specifically, SFID consists of three layers, taking spatial and frequency domain features as inputs. The outputs ${\mathbf I}^{HR}_{r}$ and ${\mathbf I}^{HR}_{up}$ contribute to the final fused image $\mathbf I$. The decoding process is illustrated in Fig.~\ref{fig:decoder}.
The complex Gabor wavelet function is defined as:
\begin{equation}
\mathcal{G}(\mathbf x)=e^{j\omega_0\mathbf x}e^{-|\upsilon_0\mathbf x|^2},
\label{eq:gabor-activation}
\end{equation}
where $\omega_0$ is the center frequency in the frequency domain, $\upsilon_0$ is a constant that is considered as the standard deviation of the Gaussian function, and $\mathbf x$ is a vector in the time (or spatial) domain. In what follows, we provide a theorem below that this Gabor wavelet activation has time-frequency tightness~\cite{timeFreqAnalysisBook}, which is helpful for the decoder's information interaction.

\noindent\textbf{Theorem 1.}
\textit{The complex Gabor wavelet activation in Eq.~\eqref{eq:gabor-activation} has the time-frequency tightness property. Moreover, from the perspective of signal spectrum analysis, this activation helps the decoder learn the optimal bandwidths.}

\noindent\textit{Proof:} The detailed proof can be found in the Supplementary.

\section{Experiments}
\noindent\textbf{Datasets:}
To evaluate the efficacy of our model, we conducted experiments using the CAVE and Harvard datasets. The CAVE dataset comprises 32 Hyperspectral Images (HSIs) with 31 spectral bands spanning from 400 nm to 700 nm at 10 nm intervals. We randomly selected 20 images for training and used the remaining 11 for testing. The Harvard dataset consists of 77 HSIs depicting indoor and outdoor scenes, covering the spectral range from 420 nm to 720 nm. We standardized the data by cropping the upper left sections of 20 Harvard images, with 10 for training and the rest for testing. Details can be found in the Supplementary.


\noindent\textbf{Implementation details:}
We implement the proposed methods FeINFN with Pytorch~\cite{NEURIPS2019_9015} on a workstation with an Intel I9 CPU and two 3090 GPUs. The optimizer is chosen as AdamW~\cite{kingma2014adam} and we use a Cosine anneal learning rate scheduler. The base channel number of the encoder is 128, that of the proposed implicit fusion function is 32 and in the decoder, the channel number is 31.

\noindent\textbf{Benchmark:}
To evaluate FeINFN's performance, we compare it with  MHIF methods on the CAVE and Harvard datasets. The bicubic-interpolated result of the upsampled LR-HSI in Tab.~\ref{table_x4} serves as our baseline. Various model-based techniques, including the CSTF-FUS ~\cite{li2018fusing}, LTTR~\cite{dian2019learning}, LTMR~\cite{dian2019hyperspectral}, and IR-TenSR~\cite{xutgrs2022} approaches, are considered. Additionally, we compare our approach with various deep learning methods, such as SSRNet~\cite{zhang2020ssr}, ResTFNet~\cite{LIU20201}, HSRNet~\cite{hu2022hyperspectral}, MoGDCN~\cite{dong2021model}, Fusformer~\cite{hu2022fusformer}, and DHIF~\cite{huang2022deep}, PSRT~\cite{deng2023psrt}, 3DT-Net~\cite{ma2023learning}, DSPNet~\cite{sun2023dual}, BDT~\cite{dengbidirectional}. 
We compare our method with other methods using different image quality metrics to validate the image fusion capability of our model, including SAM~\cite{yuhas1992discrimination}, ERGAS~\cite{wald2002data}, PSNR~\cite{5596999}, and SSIM~\cite{wang2004image}.

\noindent\textbf{Results on CAVE Dataset:}
In this section, we evaluate the effectiveness of FeINFN on the CAVE dataset and compare it with five traditional methods and some state-of-the-art deep learning-based approaches. As shown in Tab.~\ref{table_x4} on the left, our method achieves optimal performance in the tasks of $\times 4$ in all metrics. In the $\times 4$ experiment, compared to currently leading methods such as DSPNet~\cite{sun2023dual}, 3DT-Net~\cite{ma2023learning}, and BDT~\cite{dengbidirectional}, our approach demonstrates improvements in PSNR by 1.29dB/1.09dB/0.17dB, respectively. The $\times 8$ experiment is detailed in the Supplementary. 
To illustrate the advantages of our method, we provide visual comparisons in Fig.~\ref{fig:3}, including close-ups and error maps to highlight specific details. Our fusion results closely match the ground truth, achieving the best quality. In comparing error maps, the darker colors indicate closer proximity to the original image. In contrast to other excellent methods, the error maps of FeINFN distinctly exhibit superior restoration effects on details.

\noindent\textbf{Results on Harvard Dataset:}
In Tab.~\ref{table_x4}, the right columns present the comparison results of our FeINFN with other methods on the Harvard dataset at scale factors 4. Our method performs exceptionally well, with only SAM being slightly surpassed by 3DT-Net~\cite{ma2023learning} and BDT~\cite{dengbidirectional}. FeINFN exhibits significant gains in PSNR/ERGAS/SSIM metrics compared to the current state-of-the-art~\cite{dengbidirectional}, with improvements of 0.14dB/0.16/0.001, respectively. The results with a scale factor of 8 can be found in the Supplementary. As depicted in Fig.~\ref{fig:psnr-comp}, our model outperforms others, highlighting the crucial role of FeINFN's continuous representation capability in high-scale factor scenarios. To better visualize the performance gap, Fig.~\ref{fig:3} illustrates the fused images and error maps, confirming that our FeINFN maintains high fidelity in recovering the texture details of the images.

\subsection{Ablation Studies}
\noindent\textbf{Upsampling methods:}
 Implicit image representation can be seen as an advanced interpolation algorithm, offering additional spatial information and parameterized weight generation. In this section, we compare INR with other upsampling methods. We replace INR with pixel-shuffle~\cite{shi2016real} and traditional CNN interpolation methods, presenting a comparative analysis. As seen in Tab.~\ref{table_Compare1}, our approach outperforms other methods in MHIF tasks.
\begin{table}[t]
    \setlength\tabcolsep{2pt} 
	\footnotesize
        \centering
	\resizebox{7.5cm}{!}{
		\begin{tabular}{l|ccccc}
			\toprule
			Methods &PSNR($\uparrow$) &SAM($\downarrow$) &ERGAS($\downarrow$) &SSIM($\uparrow$) \\ 
            \midrule
            Bilinear  &52.23$\pm$4.40 &\ 1.92$\pm$0.60 &\ 1.03$\pm$0.86 &0.997$\pm$0.0021 \\
	      Bicubic  &52.22$\pm$4.31 &\ 1.95$\pm$0.61 &\ 1.02$\pm$0.82 &0.997$\pm$0.0021 \\
            Pixel Shuffle  &52.26$\pm$4.37 &\ {\textbf{1.90$\pm$0.59}} &\ 1.02$\pm$0.85 &0.997$\pm$0.0022 \\
            Our  &\textbf{52.47$\pm$4.10}    &\ 1.91$\pm$0.59 &\ {\textbf{0.98$\pm$0.74}} &{\textbf{0.998$\pm$0.0015}} \\
			\bottomrule
	\end{tabular}}
    \caption{Quantitative comparisons with other upsampling methods on the CAVE ($\times 4$) dataset.}
    \vspace{-0.5cm}
    \label{table_Compare1}
\end{table}
\begin{table}[!t]
    \centering
    \setlength\tabcolsep{2pt} 
	\footnotesize
	\centering
	\resizebox{7cm}{!}{
		\begin{tabular}{c|ccccc}
			\toprule
			{$\mathcal{S}$\ \ \ $\mathcal{F}$} 
            &PSNR($\uparrow$) &SAM($\downarrow$) &ERGAS($\downarrow$) &SSIM($\uparrow$) \\ 
            \midrule
			{\usym{2713}\ \ \ \usym{2717}} &52.11$\pm$4.22 &\ 1.95$\pm$0.59 &\ 1.04$\pm$0.82 &0.998$\pm$0.0017 \\
            {\usym{2717}\ \ \ \usym{2713}}&47.86$\pm$3.42 &\ 3.49$\pm$1.30 &\ 1.67$\pm$1.13 &0.995$\pm$0.0020  \\
            {\usym{2713}\ \ \ \usym{2713}}&{\textbf{52.47$\pm$4.10}}    &\ {\textbf{1.91$\pm$0.59}} &\ {\textbf{0.98$\pm$0.74}} &{\textbf{0.998$\pm$0.0015}} \\
		\bottomrule
	\end{tabular}}
    \caption{Quantitative comparisons with reduced models on the CAVE ($\times 4$) dataset. $\mathcal S$ \& $\mathcal F$ mean the domain difference.}
    \label{table_Compare2}
\end{table}
\begin{figure}
    \centering
    \includegraphics[width=7.7cm]{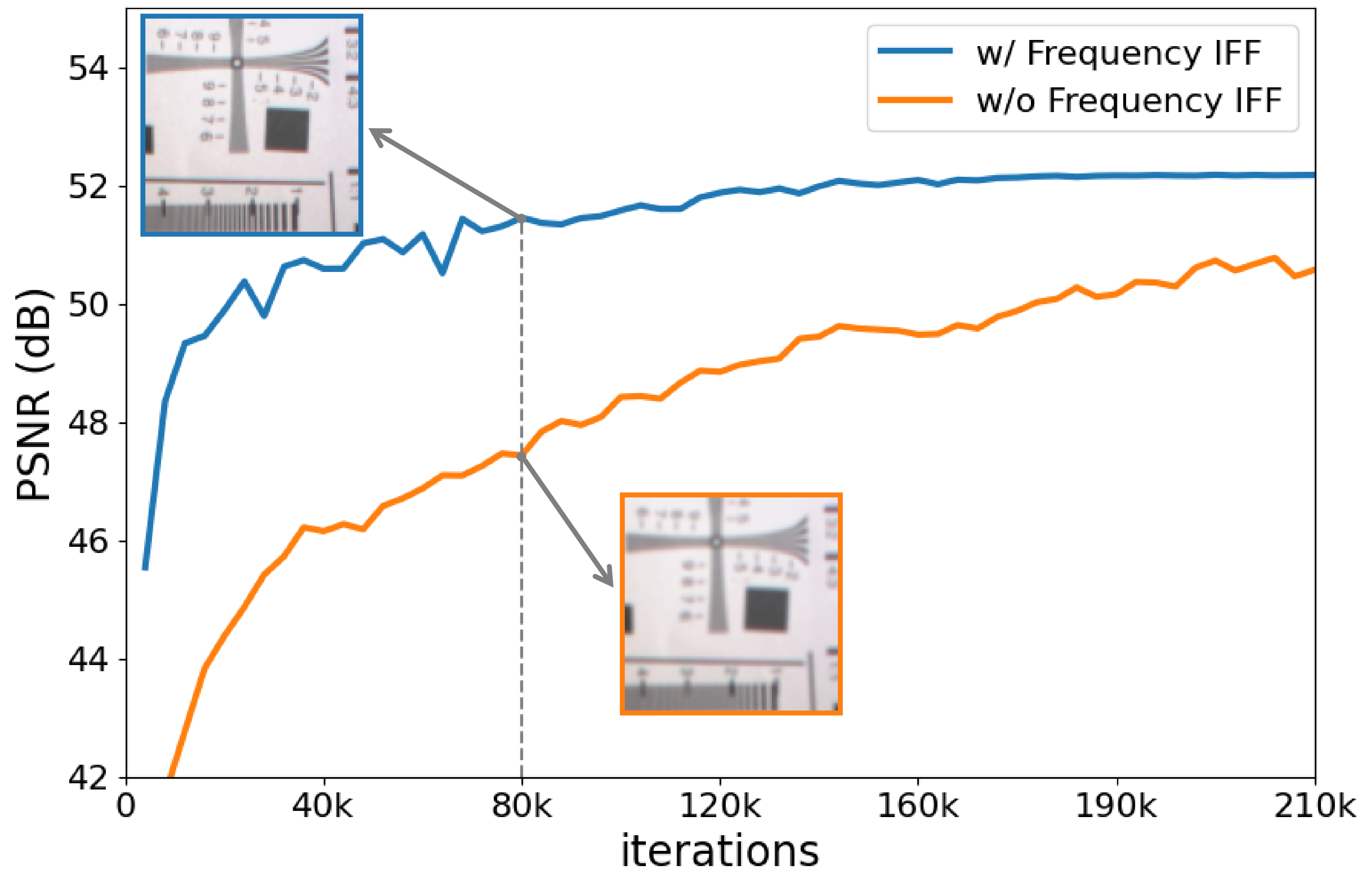}
    \vspace{-0.2cm}
    \caption{Changes in PSNR on the CAVE dataset of our FeINFN over iterations with and without the ``Fourier Domain''. The Frequency IFF can help the network learn the high-frequency details and converge faster.}
    \vspace{-0.5cm}
    \label{fig:hp}
\end{figure}

\noindent\textbf{Spatial domain and Fourier domain:}
To assess the dual-domain model's efficacy, we performed model reduction, preserving spatial and Fourier domains independently. As shown in Tab.~\ref{table_Compare2}, FeINFN excels by using both spatial and Fourier domains concurrently, underscoring the positive impact of Fourier domain integration on overall network performance.

Spectral deviation occurs during training, where the network tends to prioritize low-frequency information, capturing high-frequency details only in later stages. To validate our resolution of this issue, we remove the ``Fourier Domain" from Spa-Fre IFF, or retain it, and the corresponding training data is illustrated in Fig.~\ref{fig:hp}. Our FeINFN, which incorporates Fourier domain fusion, leads to faster PSNR convergence and overall higher efficiency. The visual comparison of high-frequency details in ``\textit{chart and stuffed toy}" from the cave dataset at 80k iterations further supports the significant improvement achieved with our results. 
\begin{table}[!t]
    \setlength\tabcolsep{2pt} 
	\footnotesize
	\centering
	\resizebox{7.7cm}{!}{
		\begin{tabular}{l|ccccc}
			\toprule
			Nonlinear &PSNR($\uparrow$) &SAM($\downarrow$) &ERGAS($\downarrow$) &SSIM($\uparrow$) \\ 
            \midrule
            ReLU  &52.03$\pm$3.84 &\ 2.00$\pm$0.59 &\ 1.02$\pm$0.74 &{\textbf{0.998$\pm$0.0013}}\\
	      GELU  &51.96$\pm$3.88 &\ 2.01$\pm$0.60 &\ 1.03$\pm$0.75 &0.998$\pm$0.0014\\
            Leaky ReLU  &51.98$\pm$3.92 &\ 2.01$\pm$0.60 &\ 1.03$\pm$0.76 &0.998$\pm$0.0014\\
            Our  &\textbf{52.47$\pm$4.10}    &\ {\textbf{1.91$\pm$0.59}} &\ {\textbf{0.98$\pm$0.74}} &0.998$\pm$0.0015\\
			\bottomrule
	\end{tabular}}
    \caption{Quantitative comparisons with different activation functions in SFID on the CAVE ($\times 4$) dataset.}
    \vspace{-0.5cm}
    \label{table_Compare3}
\end{table}

\noindent\textbf{Decoder with Different Nonlinear:}
In this section, we evaluate the impact of different activation functions in SFID, aiming to match SFIFF. Our dual-input decoder incorporates a complex Gabor wavelet activation function to facilitate the fusion of spatial and frequency domain features.
Through experiments, we replaced the Gabor wavelet activation with other activations, presenting the results in Tab.~\ref{table_Compare3}. The findings distinctly demonstrate the enhanced fusion quality achieved with the complex Gabor wavelet activation. This emphasizes the critical role of wavelet activation in promoting robust and reliable learning in SFID.
\section{Conclusion}
Inspired by the distinct behaviors of LR-HSI and HR-HSI in the Fourier domain, we introduce a novel Fourier-enhanced Implicit Neural Fusion Network (FeINFN) based on INR. Through Fourier transformation, latent features are converted into the frequency domain, allowing the modeling of frequency components to enrich high-frequency information in images. 
Additionally, we propose a spatial-frequency decoding module, achieving a unified representation of both spatial and frequency domains using a time-frequency-tight activation function.
Thanks to the unique design of our network, it outperforms state-of-the-art methods in MHIF with appealing efficiency. 
We desire that our work will inspire future research on frequency fusion-based MHIF methods.

\bibliographystyle{named}
\bibliography{ijcai24}

\end{document}